\documentclass[runningheads]{llncs}
\usepackage{graphicx}

\usepackage{tikz}
\usepackage{comment}
\usepackage{amsmath,amssymb} 
\usepackage{color}
\usepackage{siunitx}
\DeclareSIUnit\fps{fps}
\usepackage{tikz}
\usepackage[caption=false,font=normalsize,labelfont=sf,textfont=sf]{subfig}
\usepackage{ulem}

\usepackage[accsupp]{axessibility}  

\usepackage{booktabs}
\usepackage{pifont}
\newcommand{\cmark}{\ding{51}}%
\newcommand{\xmark}{\ding{55}}%
\usepackage{array}
\newcolumntype{C}[1]{>{\centering\arraybackslash}m{#1}}
\newcolumntype{L}[1]{>{\raggedright\arraybackslash}p{#1}}
\usepackage{multirow}

\begin{document}
\pagestyle{headings}
\mainmatter
\def\ECCVSubNumber{0975}  

\title{Gesture Recognition with Keypoint and Radar Stream Fusion for Automated Vehicles}

\titlerunning{Gesture Recognition with Keypoint and Radar Stream Fusion}


%
\author{Adrian Holzbock\inst{1} \and 
Nicolai Kern\inst{2} \and 
Christian Waldschmidt\inst{2} \and 
Klaus Dietmayer\inst{1} \and 
Vasileios Belagiannis\inst{3}}
\authorrunning{A. Holzbock et al.}
%
\institute{
Institute of Measurement, Control and Microtechnology, Ulm University, Albert-Einstein-Allee 41, 89081 Ulm, Germany\\
\email{first\_name.last\_name@uni-ulm.de}
\and
Institute of Microwave Engineering, Ulm University, Albert-Einstein-Allee 41, 89081 Ulm, Germany\\
\email{first\_name.last\_name@uni-ulm.de}
\and
Department of Simulation and Graphics, Otto von Guericke University Magdeburg, Universitätsplatz 2, 39106 Magdeburg, Germany\\
\email{first\_name.last\_name@ovgu.de}
}
\maketitle

\setcounter{footnote}{0}

\begin{abstract}
We present a joint camera and radar approach to enable autonomous vehicles to understand and react to human gestures in everyday traffic. Initially, we process the radar data with a PointNet followed by a spatio-temporal multilayer perceptron (stMLP). Independently, the human body pose is extracted from the camera frame and processed with a separate stMLP network. We propose a fusion neural network for both modalities, including an auxiliary loss for each modality. In our experiments with a collected dataset, we show the advantages of gesture recognition with two modalities. Motivated by adverse weather conditions, we also demonstrate promising performance when one of the sensors lacks functionality. 

\end{abstract}

\section{Introduction}
The safe interaction of traffic participants in urban environments is based on different rules like signs or the right of way. Besides static regulations, more dynamic ones like gestures are possible. For example, a police officer manages the traffic~\cite{wiederer2020traffic} by hand gestures, or a pedestrian waves a car through at a crosswalk~\cite{rasouli2019autonomous}. Although the driver intuitively knows the meaning of human gestures, an autonomous vehicle cannot interpret them. To safely integrate the autonomous vehicle into urban traffic, it is essential to understand human gestures. 

To detect human gestures, many related approaches only rely on camera data~\cite{baek2022traffic,pham2021efficient}. Camera-based gesture recognition is not always reliable, for instance, due to the weather sensitivity of the camera sensor~\cite{9011192}. One way to mitigate the drawbacks of camera-based approaches is the augmentation of the gesture recognition system by non-optical sensor types. A sensor with less sensitivity to environmental conditions is a radar sensor~\cite{qian2021robust} that has been shown to be suited for gesture recognition. Furthermore, radar sensors are not limited to detecting fine-grained hand gestures but can also predict whole body gestures~\cite{kern2022pointnet+}. Hence, radar sensors are a promising candidate to complement optical sensors for more reliable gesture recognition in automotive scenarios. This has been demonstrated for gesture recognition with camera-radar fusion in close proximity to the sensors, as presented in~\cite{molchanov2015multi}. Furthermore, depending not only on one sensor can increase the system's reliability in case of a sensor failure. In contrast to the prior work, we propose a method for whole-body gesture recognition at larger distances and with a camera-radar fusion approach.

We present a two-stream neural network to realizing camera and radar fusion. Our approach extracts independently a representation for each modality and then fuses them with an additional network module. In the first stream, the features of the unordered radar targets are extracted with a PointNet~\cite{qi2017pointnet} and further prepared for fusion with an spatio-temporal multilayer perceptron (stMLP)~\cite{holzbock2022spatio}. The second stream uses only a stMLP to process the keypoint data. The extracted features of each stream are fused for the classification in an additional stMLP. We train the proposed network with an auxiliary loss function for each modality to improve the feature extraction through additional feedback. For training, we use data containing radar targets of three chirp sequence (CS) radar sensors and human keypoints extracted from camera images. The eight different gestures are presented in Fig.~\ref{fig:gesture_set} and represent common traffic gestures. The performance of our approach is tested in a cross-subject evaluation. We improve performance by 3.8 percentage points compared to the keypoint-only setting and 8.3 percentage points compared to the radar-only setting. Furthermore, we can show robustness against the failure of one sensor.

Overall, we propose a two-stream neural network architecture to fuse keypoint and radar data for gesture recognition. To process the temporal context in the model, we do not rely on recurrent models~\cite{9196702} but instead propose the stMLP fusion. In the training of our two-stream model, we introduce an auxiliary loss for each modality. To the best of our knowledge, we are the first to fuse radar and keypoint data for gesture recognition in autonomous driving. 

\newcommand\imgw{0.173}
\begin{figure}[!ht]
\centering

\subfloat[]{
\centering
\begin{minipage}[b]{\imgw\linewidth}
\centering
\includegraphics[width=1.0\linewidth]{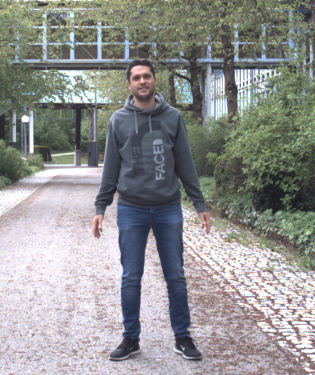}%
\hfill
\vspace{1em}
\includegraphics[width=1.0\linewidth]{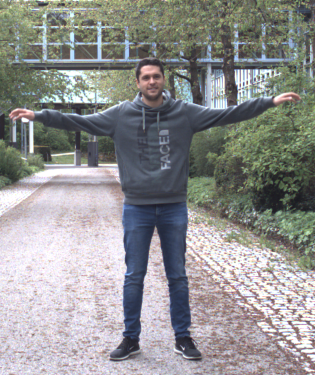}%
\end{minipage}%
}
\quad
\subfloat[]{
\begin{minipage}[b]{\imgw\linewidth}
\centering
\includegraphics[width=1.0\linewidth]{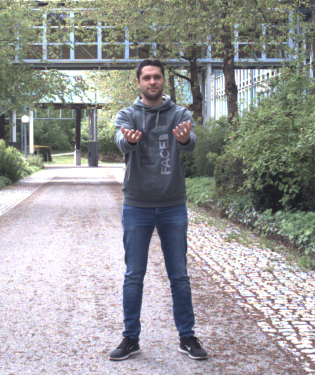}%
\hfill
\vspace{1em}
\includegraphics[width=1.0\linewidth]{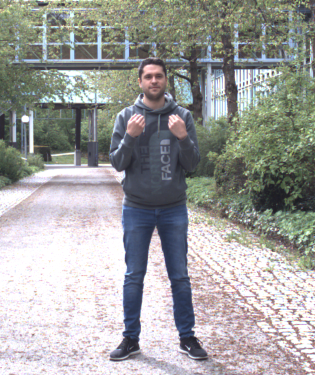}%
\end{minipage}%
}
\quad
\subfloat[]{
\begin{minipage}[b]{\imgw\linewidth}
\centering
\includegraphics[width=1.0\linewidth]{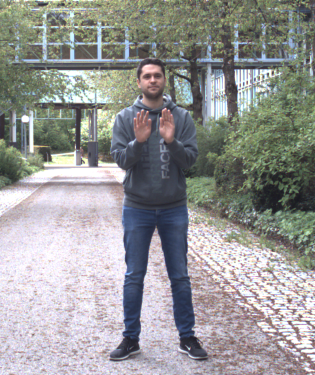}%
\hfill
\vspace{1em}
\includegraphics[width=1.0\linewidth]{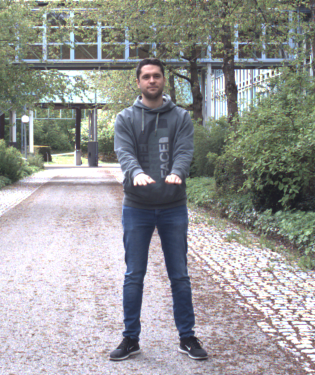}%
\end{minipage}%
}
\quad
\subfloat[]{
\begin{minipage}[b]{\imgw\linewidth}
\centering
\includegraphics[width=1.0\linewidth]{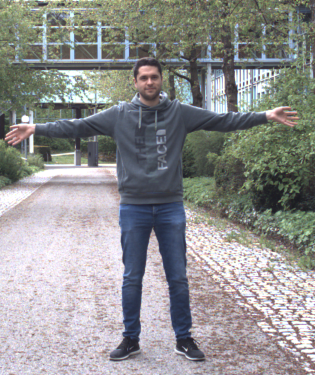}%
\hfill
\vspace{1em}
\includegraphics[width=1.0\linewidth]{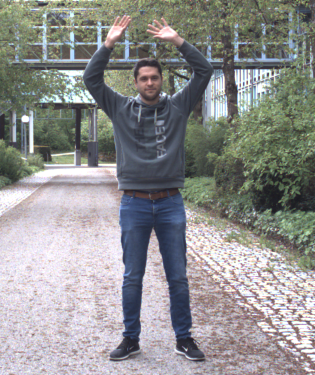}%
\end{minipage}%
}
\quad
\par\smallskip
\subfloat[]{
\begin{minipage}[b]{\imgw\linewidth}
\centering
\includegraphics[width=1.0\linewidth]{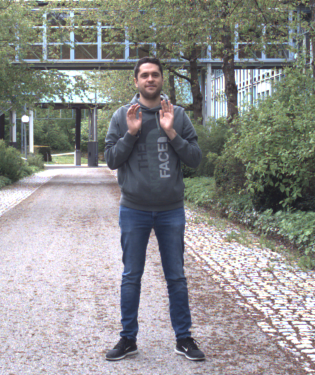}%
\hfill
\vspace{1em}
\includegraphics[width=1.0\linewidth]{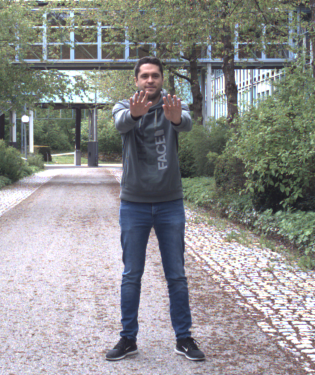}%
\end{minipage}%
}
\quad
\subfloat[]{
\begin{minipage}[b]{\imgw\linewidth}
\centering
\includegraphics[width=1.0\linewidth]{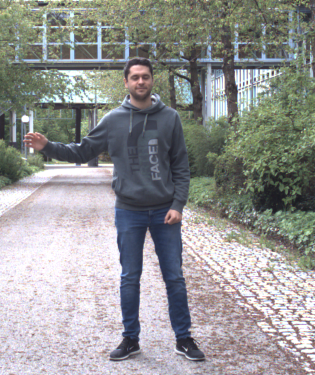}%
\hfill
\vspace{1em}
\includegraphics[width=1.0\linewidth]{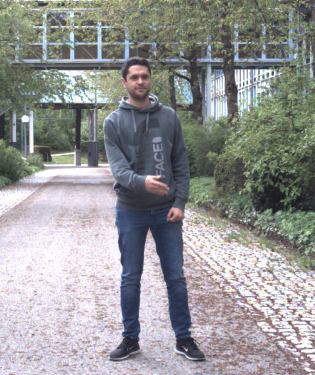}%
\end{minipage}%
}
\quad
\subfloat[]{
\begin{minipage}[b]{\imgw\linewidth}
\centering
\includegraphics[width=1.0\linewidth]{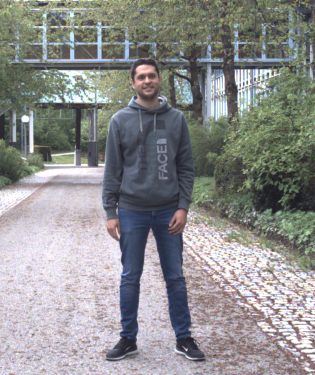}%
\hfill
\vspace{1em}
\includegraphics[width=1.0\linewidth]{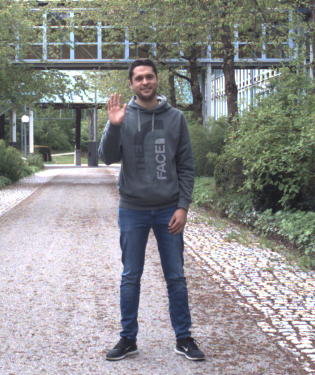}%
\end{minipage}%
}
\quad
\subfloat[]{
\begin{minipage}[b]{\imgw\linewidth}
\centering
\includegraphics[width=1.0\linewidth]{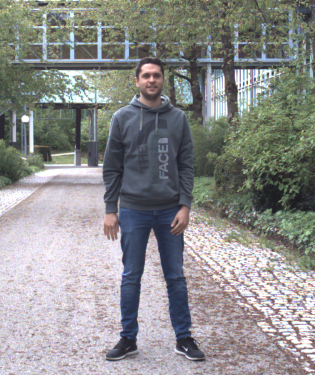}%
\hfill
\vspace{1em}
\includegraphics[width=1.0\linewidth]{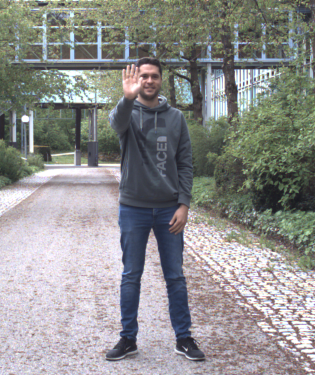}%
\end{minipage}%
}
\caption{Visualization of the characteristic poses of the eight gestures. (a)~Fly. (b)~Come closer. (c)~Slow down. (d)~Wave. (e)~Push away. (f)~Wave through. (g)~Stop. (h)~Thank you.}
\label{fig:gesture_set}
\end{figure}

\section{Related Work}
In the following, we discuss other methods regarding gesture recognition in general as well as in autonomous driving. We give an overview of approaches relying only on keypoint or radar data. Besides the single sensor gesture recognition, we present methods using a combination of different sensors.

\paragraph{\textbf{Gesture Recognition with Camera Data}}
Gesture recognition is applied to react to humans outside the vehicle~\cite{xu2021action,pham2021efficient} and to the passenger's desires~\cite{holzbock2022spatio,wharton2021coarse}. For gesture recognition of police officers, related work processes camera image snippets~\cite{baek2022traffic} directly. Due to the advances in human body pose estimation~\cite{9667074,belagiannis2014holistic}, related approaches extract the body skeleton data from the images and perform gesture recognition on it~\cite{wang2021simple,pham2021efficient,mishra2021authorized}. For processing skeletons to predict the gesture, recurrent neural networks~\cite{wang2021simple} or convolutional neural networks~\cite{pham2021efficient} are applied. Besides the police officer gesture recognition, the actions of other human traffic participants like cyclists~\cite{xu2021action} or pedestrians~\cite{geng2020using} are also analyzed in literature. Similar to pedestrian gesture recognition is the pedestrian intention prediction~\cite{quintero2017pedestrian,abughalieh2020predicting}, where the pedestrian's intention to cross the street should be recognized. Changing the view to the interior of the car, there are approaches to recognize the driver's activities in order to check if the driver is focused on the traffic. For this purpose, methods like attention-based neural networks~\cite{wharton2021coarse} or models only built on multi-layer perceptrons~\cite{holzbock2022spatio} are developed. Compared to our work, lighting and weather conditions influence the performance of camera-based approaches. Due to the fusion with the radar sensor data, our method mitigate these factors.

\paragraph{\textbf{Gesture Recognition with Radar Data}}
Besides radar sensors' insensitivity to adverse weather and lighting, they also evoke fewer privacy issues than cameras. As a result, radar-based gesture recognition has received increased attention in recent years, with research efforts mainly devoted to human-machine interaction in the consumer electronics area \cite{lien2016soli}. For the control of devices with hand gestures, gesture recognition algorithms based on a wide range of neural networks have been proposed, involving, e.g., 2D-CNNs \cite{kim2016hand}, 2D-CNNs with LSTMs \cite{wang2016interacting}, or 3D-CNNs with LSTM \cite{zhang2018latern}. These approaches exploit spectral information in the form of micro-Doppler spectrograms \cite{kim2016hand} or range-Doppler spectra \cite{wang2016interacting}, but it is also possible to distinguish between gestures \cite{kern2022pointnet+} and activities \cite{singh2019radhar} using radar point clouds. The latter are a more compact representation of the radar observations and are obtained by finding valid targets in the radar data. Point clouds facilitate the application of geometrical transformations \cite{singh2019radhar} as well as the inclusion of additional information \cite{kern2022pointnet+}. While most research considers small-scale gesture recognition close to the radar sensor, reliable macro gesture recognition at larger distances has been also shown to be feasible with radar sensors for applications such as smart homes \cite{ninos2021real} or traffic scenarios \cite{kern2020robust,kern2022pointnet+}. While radar-only gesture recognition has shown promising results, augmenting it by camera data can further improve classification accuracy, as demonstrated in this paper. This is particularly important in safety-critical applications such as autonomous driving. 

\paragraph{\textbf{Gesture Recognition with Sensor Fusion}}
By combining data of multiple sensors, sensor fusion approaches can overcome the drawbacks of the individual sensor types, like the environmental condition reliance of the camera or its missing depth information. Sensor fusion has been applied e.g. for gesture-based human machine interaction in vehicles, where touchless control can increase safety ~\cite{molchanov2015hand}. Besides the image information for gesture recognition, the depth data contains essential knowledge. Therefore, other methods fuse camera data with the data of a depth sensor~\cite{ohn2014hand} and process the data with a 3D convolutional neural network~\cite{molchanov2015hand,zengeler2018hand}. Molchanov et al.~\cite{molchanov2015multi} run a short-range radar sensor next to the RGB camera and depth camera and process the recorded data with a 3D convolutional neural network for a more robust gesture recognition. Another approach fuses the short-range radar data with infrared sensor data instead of RGB images~\cite{skaria2021radar}. Current fusion approaches are not only limited to short ranges and defined environments but can also be applied in more open scenarios like surveillance~\cite{de2021classification} or smart home applications~\cite{vandersmissen2020indoor}. Here, camera images and radar data, converted to images, are used to classify the gesture, while we use the skeletons extracted from the images and the radar targets described as unstructured point clouds. Moreover, contrary to~\cite{vandersmissen2020indoor}, our fusion approach enhances the gesture recognition accuracy not only in cases where one modality is impaired but also in normal operation.

\section{Fusion Method}
We present a fusion technique for radar and keypoint data for robust gesture recognition. To this end, we develop a neural network ${\mathbf{\hat{y}} = f((\mathbf{x}_R, \mathbf{x}_K), \theta)}$ defined by its parameters $\theta$. The output prediction $\mathbf{\hat{y}}$ is defined as one-hot vector ${\mathbf{\hat{y}}_i \in \{0,1\}^{C}}$, such that $\sum_{c=1}^{C}\mathbf{\hat{y}}_i(c)=1$ for a C-category classification problem. The radar input data is represented by ${\mathbf{x}_R \in \mathbb{R}^{T \times 5 \times 300}}$, where T is the number of time steps and 300 is the number of sampled radar targets in each time step, each of which is described by 5 target parameters. For the keypoint stream, 17 2D keypoints are extracted and flattened in each time step, such that the keypoint data is described by ${\mathbf{x}_K \in \mathbb{R}^{T \times 34}}$. We use the ground truth class label $\mathbf{y}$ for the training of $f(.)$ to calculate the loss of the model prediction $\mathbf{\hat{y}}$. In the following, we first present the architecture of our neural network and afterward show the training procedure.

\begin{figure}[!ht]
    \centering
    \includegraphics[width=0.53\textwidth]{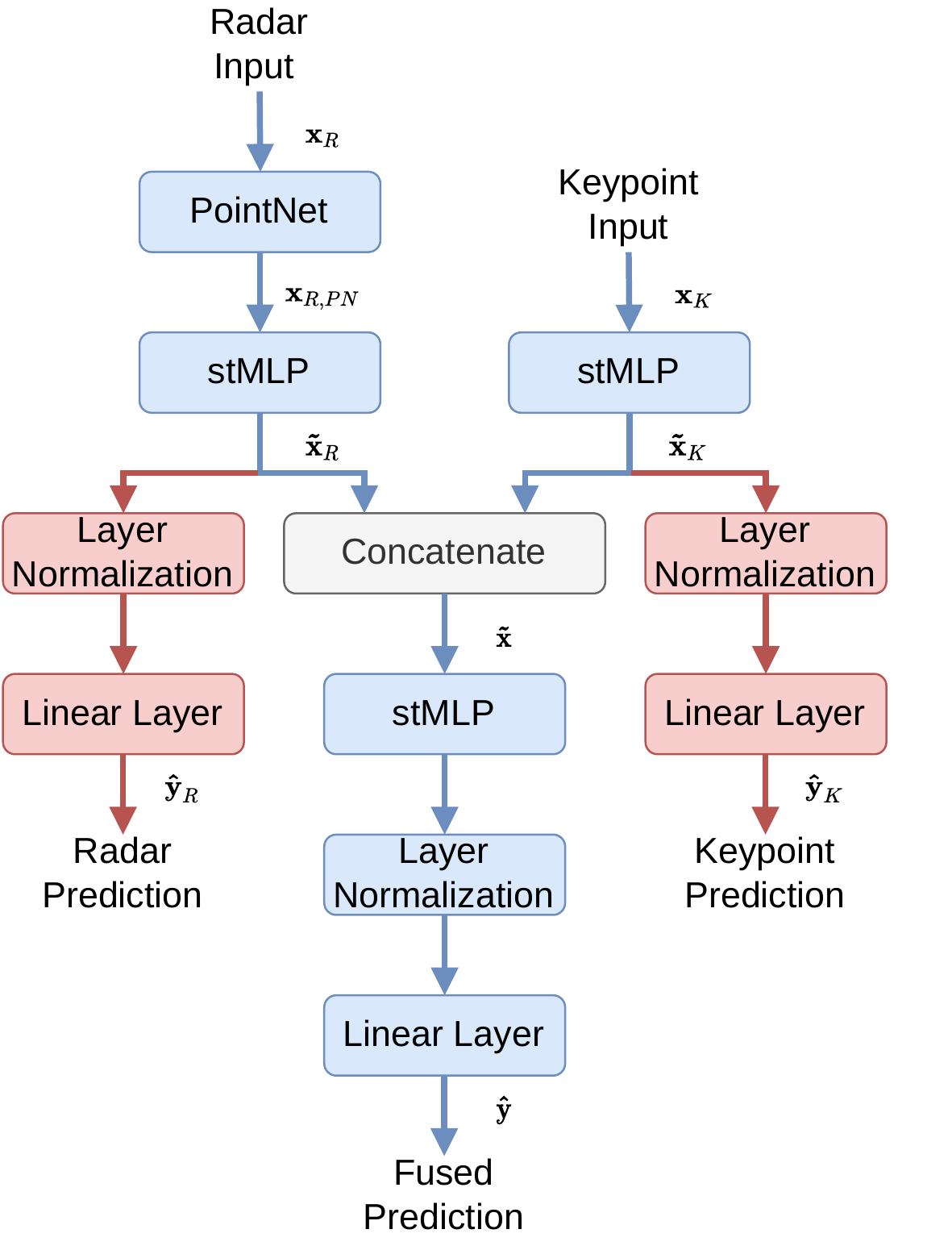}
    \caption{Architecture of the proposed neural network for gesture recognition with radar and keypoint data. The blue layers correspond to the inference neural network while the red layers are only used to compute the auxiliary losses.}
    \label{fig:overview_model}
\end{figure}

\subsection{Neural Network Architecture}
\label{subsec:architecture}
The neural network inputs are the radar data $\mathbf{x}_R$ and the keypoint data $\mathbf{x}_K$. The network consists of two different streams, one for each modality, to extract the features of the different modalities. The information of the two streams is concatenated and fed to a joint network which does the fusion and the gesture classification. Additionally, auxiliary outputs are added to the model for the training procedure. An overview of the proposed network architecture is given in Fig.~\ref{fig:overview_model}. 

The feature extraction from the unordered radar data ${\mathbf{x}_R}$ is done with a PointNet~\cite{qi2017pointnet} that extracts the features for each time step ${t \in \{1, 2, \dots, T\}}$. The features of the time steps are concatenated to one feature tensor ${\mathbf{x}_{R,PN} \in \mathbb{R}^{T \times 512}}$. The PointNet does not process the data in the temporal dimension but only in the point dimension. For the temporal processing, we use the stMLP model~\cite{holzbock2022spatio}, which replaces a standard method for temporal data processing, e.g. a long short-term memory model (LSTMs)~\cite{hochreiter1997long}. Unlike LSTMs, the stMLP is based solely on multilayer perceptrons (MPLs) and does not have any recurrent parts. The stMLP  processes the radar data $\mathbf{x}_{R,PN}$ extracted with the PointNet in the temporal and feature dimensions and outputs mixed radar features ${\mathbf{\tilde{x}}_{R} \in \mathbb{R}^{T \times H/2}}$, where $H$ is the hidden dimension. Due to the ordered structure of the keypoint data $\mathbf{x}_K$, the mixed keypoint features ${\mathbf{\tilde{x}}_K \in \mathbb{R}^{T \times H/2}}$ are extracted only with an stMLP model, which processes the extracted keypoint features in both dimensions, namely the spatial and the temporal dimension. The extracted features of both modalities, $\mathbf{\tilde{x}}_R$ and $\mathbf{\tilde{x}}_K$, are concatenated to a single feature tensor ${\mathbf{\tilde{x}} \in \mathbb{R}^{T \times H}}$ and fed into another stMLP model. This model performs a spatial and temporal fusion of the radar and keypoint features to produce a meaningful representation for the gesture classification step. The gesture classification is performed by a Layer Normalization~\cite{ba2016layer} and a single linear layer for each time step.

Additionally, we add for each modality an auxiliary output during training~\cite{szegedy2015going}. The auxiliary output of both modalities is built on a Layer Normalization and a linear layer for the auxiliary classification. In Fig.~\ref{fig:overview_model}, the parts of the auxiliary outputs and the corresponding layers are drawn in red. For the gesture prediction, each auxiliary output only uses one modality and no information from the other. This means that the auxiliary radar output $\mathbf{\hat{y}}_R$ gets the extracted radar features $\mathbf{\tilde{x}}_R$ for the prediction and the auxiliary keypoint output $\mathbf{\hat{y}}_K$ the keypoint data features $\mathbf{\tilde{x}}_K$. For the model, the auxiliary outputs give additional specialized feedback for updating the network parameters in the radar and the keypoint stream. After training, the layers of the auxiliary outputs can be removed.

\subsection{Model Training}
For the training of the neural network introduced in Sec.~\ref{subsec:architecture} we utilize the data presented in Sec.~\ref{subsec:dataset}. During training, we use the fused output of the model $\mathbf{\hat{y}}$ and the auxiliary outputs derived from the radar $\mathbf{\hat{y}}_R$ and keypoint data $\mathbf{\hat{y}}_K$. For each output, we calculate the cross entropy loss $\mathcal{L}_{CE}$, which can be formulated as 
\begin{equation}
    \mathcal{L}_{CE} = - \sum^{C}_{c=0}y_c \log (\hat{y}_c),
\end{equation}
where $y$ is the ground truth label, $\hat{y}$ the network prediction, and $C$ the number of gesture classes. We use a weighted sum of the different sub-losses to get the overall loss $\mathcal{L}$ which can be expressed by
\begin{equation}
\label{eq:loss}
    \mathcal{L} = \mathcal{L}_F + \mu * ( \mathcal{L}_R + \mathcal{L}_K ).
\end{equation}
In the overall loss, $\mathcal{L}_F$ is the loss of the fused output, $\mathcal{L}_R$ of the auxiliary radar output, and $\mathcal{L}_K$ of the auxiliary keypoint output. The auxiliary losses are weighted with the auxiliary loss weight $\mu$.

\section{Results}
We evaluate our approach in two different settings. First, we test our approach with two modalities, assuming no issue with the sensors and getting data from both. Second, we use only one modality to evaluate the model. Only having one modality can be motivated by adverse weather conditions, e.g. the camera sensor is completely covered by snow or technical problems with one sensor.

\subsection{Dataset}
\label{subsec:dataset}
For the development of robust gesture recognition with the fusion of radar and keypoint data in the context of autonomous driving, we need a dataset that contains both modalities captured synchronously and at sufficient ranges. Consequently, the custom traffic gesture dataset introduced in~\cite{kern2022pointnet+} with both camera and radar data is used. Following, we specify the setup and the data processing of the dataset.

\begin{figure}[!ht]
    \centering
    \includegraphics[width=0.7\linewidth]{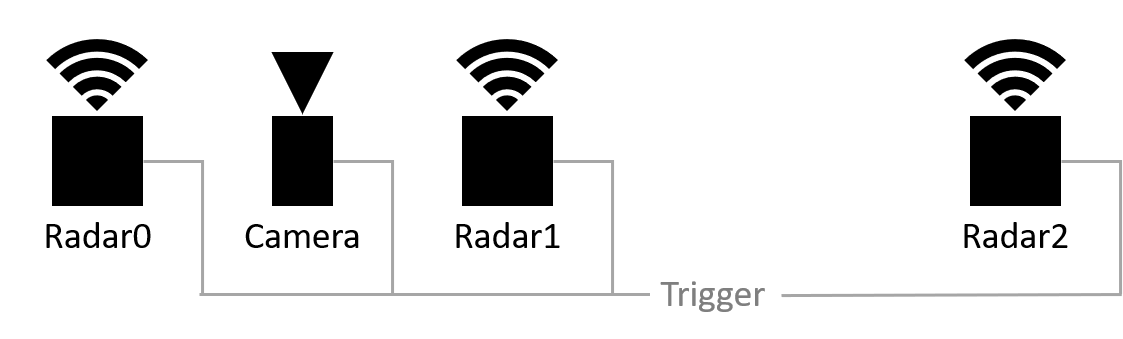}
    \caption{Measurement system consisting of the RGB camera and three CS radar sensors. All sensors receive a common trigger signal for time synchronization.}
    \label{fig:sensor_setup}
\end{figure}

\paragraph{\textbf{Setup}}
The gesture dataset comprises measurements of eight different gestures shown in Fig.~\ref{fig:gesture_set} for 35 participants. The measurements are conducted on a small street on the campus of Ulm University as well as inside a large hall resembling a car park. For each participant, data recording is repeated multiple times under different orientations, and the measurements are labeled by means of the camera data. For the measurements, a setup consisting of three chirp sequence (CS) radar sensors and a RGB camera as illustrated by the sketch in Fig.~\ref{fig:sensor_setup} are used. The radar sensors operate in the automotive band at \SI{77}{\giga\hertz}. Each sensor has a range and velocity resolution of \SI{4.5}{\centi\meter} and \SI{10.7}{\centi\meter\per\second}, respectively, and eight receive channels for azimuth angle estimation. The camera in the setup has a resolution of 1240$\times$1028, and keypoints obtained from the camera serve as input to the keypoint stream of the proposed gesture recognition algorithm. All sensors are mounted on a rail and synchronized with a common trigger signal with \SI{30}{\fps}.  

\paragraph{\textbf{Data Processing}}
From the gesture recordings, per-frame radar point clouds and 2D skeletal keypoints are computed. The radar responses recorded by the sensors are processed sensor-wise to obtain range-Doppler maps \cite{Winkler.2007}. The Ordered Statistics CFAR algorithm \cite{Rohling.1983} is applied to extract valid targets and thereby compress the information in the range-Doppler maps. For each radar target, its azimuth angle is estimated by digital beamforming \cite{Vasanelli.2020}. The target parameters are normalized by the radar sensors' unambiguous ranges. Each target in the resulting target lists is described by its range, velocity, azimuth angle, its reflected power in \si{\decibel}, and the index ${i_n \in {0,1,2}}$ of the radar sensor that detected it. Since the number of detected targets varies from frame to frame, target lists with a constant number $N_R$ of targets are sampled randomly for frames whose target count exceeds $N_R$. Contrary, for frames where the number of targets is less than $N_R$, zero-padding is applied to fill the target lists. The target lists of the radar sensors are stacked, such that the final target list contains ${3N_R}$ targets. After repeating the processing for all $T_M$ frames in the measurement, the radar observations are described by input data of shape ${\mathbf{x}_R \in \mathbb{R}^{T_M \times 5 \times 300}}$, when setting $N_R$ to 100.

The camera data is processed by Detectron2~\cite{wu2019detectron2} to extract 17 keypoints in the COCO keypoint format~\cite{Lin.2014}. The extracted keypoints are normalized with the image width and height to restrict the keypoint values for the neural network to the range between 0 and 1. The keypoints' 2D pixel positions over the frames are summarized in the camera observation tensor ${\mathbf{o}_K \in \mathbb{R}^{T_M \times 2 \times 17}}$, which is flattened to the keypoint data ${\mathbf{x}_K \in \mathbb{R}^{T_M \times 34}}$. After the signal processing, the radar and camera data of the measurements are downsampled to \SI{15}{\fps} and segmented into smaller snippets with ${T=30}$ time steps each, corresponding to \SI{2}{\second} of observation. Finally, 15700 samples are available for training the gesture recognition model. In the cross-subject evaluation, we use the data of 7 subjects for testing and the data of the remaining subjects as training and validation data. We train the model 5 times, using different subjects for testing in each run and average over the 5 runs for the final performance.

\subsection{Experimental Setup}
\label{subsec:setup}
The neural network is implemented with the PyTorch~\cite{NEURIPS2019_9015} framework, and we use PointNet\footnote{https://github.com/fxia22/pointnet.pytorch} and stMLP\footnote{https://github.com/holzbock/st\_mlp} as a base for our new fusion architecture. In the PointNet, we apply the feature transformation and deactivate the input transformation. In each stMLP structure we use 4 mixer blocks and set the stMLP hyperparameters as follows: the hidden input dimension to 256, the hidden spatial-mixing dimension to 64, and the hidden temporal-mixing dimension to 256. The loss is calculated with the function defined in Eq.~\ref{eq:loss}, where we set the auxiliary weight $\mu$ to 0.5. We train our model for 70 epochs with a batch size of 32 and calculate the gradients with the SGD optimizer that uses a learning rate of 0.003, a momentum of 0.95, and a weight decay of 0.001. To get the best training result, we check the performance during the training on the validation set and use the best validation epoch for testing on the test set. The model performance is measured with the accuracy metric. To prepare the neural network for missing input data, we skip in 30\% of the training samples the radar or the keypoint data, which we call skipped-modality training (SM-training). When reporting the results with a single-modality model, we remove the layers for the other modality.

\subsection{Results}
During testing with optimal data we assume that we get the keypoint data from the camera sensor and the targets from the radar sensors, i.e., we have no samples in the train and test set that only contain one modality's data. The results are shown in Tab.~\ref{table:correct_data}, where we compare with a model using an LSTM instead of an stMLP for temporal processing. In the \textit{Single Modality} part of the table, we show the performance of our architecture trained and tested on only one modality. In these cases, the layers belonging to the other modality are removed. As it can be seen, the LSTM and the stMLP model are on par when training only with the radar data, and the stMLP model is better than the LSTM model trained only with the keypoint data. 

In the \textit{Fusion} part of Tab.~\ref{table:correct_data}, we show the fusion performance of our architecture. We first train the stMLP and the LSTM fusion with all the training data and then apply SM-training with a ratio of 30\%. The SM-training means that we skip the radar or the keypoint data in 30\% of the training samples. The stMLP fusion performs better in both cases compared to the LSTM fusion. Furthermore, skipping randomly one modality in 30\% of the training data slightly benefits the test performance. Overall, the stMLP architecture performs better in the fusion of the radar and the keypoint data than the LSTM architecture. Compared to the single-modality model, the fusion improves the performance in both architectures (LSTM and stMLP) by over 4 percentage points. This shows that the fusion of the keypoint and radar data for gesture recognition benefits the performance compared to single-modality gesture recognition.

\setlength{\tabcolsep}{4pt}
\begin{table}[ht]
\begin{center}
\caption{Performance of our model trained with both modalities. \textit{Single Modality} is a model that only contains layers for one modality and is trained and tested only with this modality. The \textit{Fusion} rows use our proposed model. \textit{SM-training} means that we skip one modality in 30\% of the training samples during the training.}
\label{table:correct_data}
\begin{tabular}{lC{1.5cm}C{1.4cm}C{1.4cm}C{1.4cm}C{1.4cm}C{1.4cm}}
\toprule
\multirow{2}{*}{Model} & Temporal & \multicolumn{2}{c}{Test data} & SM- & Accuracy \\
& Processing & Radar & Keypoint & training & in \% \\
\midrule
\multirow{4}{*}{Single Modality} & LSTM & \xmark & \cmark & \xmark & 86.2 \\ 
& stMLP & \xmark & \cmark & \xmark & 89.9 \\
& LSTM & \cmark & \xmark & \xmark & 85.7 \\
& stMLP & \cmark & \xmark & \xmark & 85.4 \\
\midrule
\multirow{4}{*}{Fusion} & LSTM & \cmark & \cmark & \xmark & 90.2 \\
& stMLP & \cmark & \cmark & \xmark & 93.7 \\
& LSTM & \cmark & \cmark & \cmark & 90.7 \\
& stMLP & \cmark & \cmark & \cmark & 93.8 \\
\bottomrule
\end{tabular}
\end{center}
\end{table}
\setlength{\tabcolsep}{1.4pt}

\subsection{Results with Single Modality}
\label{subsec:corrupted_data}
When we test our architecture with single modality, during testing only the radar or the keypoint data are fed into the model, but not both. This can be compared with one fully corrupted sensor due to adverse weather or a technical problem. Contrary, during training we either use all training data (no SM-training) or set the SM-training to a ratio of 30\%, such that in 30\% of samples one modality is skipped. In the first part of Tab.~\ref{table:corrupted_data}, results are shown for the model trained without the skipped modalities, which means that the model has not learned to handle single-modality samples. Despite the missing modality in the evaluation, the model can still classify the gestures, but with a decreased accuracy. Comparing the LSTM and stMLP variants, the LSTM variant performs better with only the radar data and the stMLP with only the keypoint data. 

In the second part of Tab.~\ref{table:corrupted_data}, we show the performance of our model trained with skipped modalities in 30\% of the training samples (SM-training is 30\%). When the model is trained with the skipped modalities, it learns better to handle missing modalities. This results in a better performance in the single-modality evaluation, and we can improve the accuracy by a minimum of 3 percentage points compared to the model trained without the skipped modalities. Overall, the fusion can improve the reliability of gesture recognition in cases where one sensor fails e.g. due to adverse weather conditions or a technical sensor failure.

\setlength{\tabcolsep}{4pt}
\begin{table}[ht]
\begin{center}
\caption{Performance of our model with single-modality data. The \textit{Fusion} rows use our proposed model. \textit{SM-training} means that we skip one modality in 30\% of the training samples during the training.}
\label{table:corrupted_data}
\begin{tabular}{lC{1.5cm}C{1.4cm}C{1.4cm}C{1.4cm}C{1.4cm}C{1.4cm}}
\toprule
\multirow{2}{*}{Model} & Temporal & \multicolumn{2}{c}{Test data} & SM- & Accuracy \\
& Processing & Radar & Keypoint & training & in \% \\
\midrule
\multirow{4}{*}{Fusion} & LSTM & \xmark & \cmark & \xmark & 69.7 \\ 
& LSTM & \cmark & \xmark & \xmark & 82.6 \\
& stMLP & \xmark & \cmark & \xmark & 87.7 \\
& stMLP & \cmark & \xmark & \xmark & 78.3 \\
\midrule
\multirow{4}{*}{Fusion} & LSTM & \xmark & \cmark & \cmark & 80.8 \\
& LSTM & \cmark & \xmark & \cmark & 85.0 \\
& stMLP & \xmark & \cmark & \cmark & 91.2 \\
& stMLP & \cmark & \xmark & \cmark & 83.0 \\
\bottomrule
\end{tabular}
\end{center}
\end{table}
\setlength{\tabcolsep}{1.4pt}

\section{Ablation Studies}
The performance of our model depends on different influence factors, which we evaluate in the ablation studies. As standard setting, we choose the hyperparameters defined in Sec.~\ref{subsec:setup} and change the ratio of the SM-training and the auxiliary loss weight during the ablations.

\subsection{Amount of Single-Modality Training Samples}
The training with missing data teaches the model to handle singe modality input data, as shown in Sec.~\ref{subsec:corrupted_data}. In our standard setting for the SM-training, we randomly skip one modality in 30\% of the training samples. In this ablation study, the SM-training ratio ranges from 0\% to 60\% during training and shows the influence of this parameter on the performance.

We test the model with different SM-training ratios with the test data containing both modalities and show the result in the \textit{Fusion} row of Tab.~\ref{table:ablation_sm}. Additionally, we test with the data of only one modality and present the results in the \textit{Only Keypoints} and \textit{Only Radar} row. The experiment shows that the SM-training does not significantly influence the performance when testing with both modalities. This is explainable because in the testing set no samples have lacking modalities. Testing with only one modality, the amount of skipped-modality samples during the training influences the accuracy. Here, the performance increases until the SM-training ratio reaches 30\% and then stays constant. In this case, the amount of skipped-modality samples during training influences the adaptation of the model to single-modality samples.

\setlength{\tabcolsep}{4pt}
\begin{table}[ht]
\begin{center}
\caption{Influence of the SM-training ratio on the overall accuracy.}
\label{table:ablation_sm}
\begin{tabular}{lC{.8cm}C{.8cm}C{.8cm}C{.8cm}C{.8cm}C{.8cm}C{.8cm}}
\toprule
Skip modality in \% & 0 & 10 & 20 & 30 & 40 & 50 & 60 \\
\midrule
Fusion & 93.7 & 92.8 & 92.1 & 93.8 & 94.1 & 93.2 & 93.6 \\
Only Keypoints & 87.7 & 89.9 & 89.8 & 91.2 & 91.4 & 91.0 & 91.5 \\
Only Radar & 78.3 & 81.7 & 81.1 & 83.0 & 83.0 & 80.7 & 83.2 \\
\bottomrule
\end{tabular}
\end{center}
\end{table}
\setlength{\tabcolsep}{1.4pt}

\subsection{Loss Function}
Besides the ratio of the SM-training, the auxiliary loss weight $\mu$ is an essential parameter in training. In our standard evaluation, we set $\mu$ to 0.5, while we vary the loss weight in this ablation study from 0 to 3.

The results of the different $\mu$ are shown in Tab.~\ref{table:ablation_lw}. In the first row, we deliver the results when testing with both modalities (\textit{Fusion}). In the \textit{Only Keypoints} and \textit{Only Radar} row, we present the performance when evaluating only one modality. As we can see in Tab.~\ref{table:ablation_lw}, the fusion results behave equally to the single-modality results. For the Fusion as well as for the single modality, a higher $\mu$ increases the performance, and the best accuracy is reached with $\mu = 0.8$. Further increasing $\mu$ leads to a decreasing performance in gesture recognition. Comparing the results of the best $\mu = 0.8$ with the model without the auxiliary loss ($\mu = 0.0$) shows that with the auxiliary loss, the fusion performance stays equal but the single-modality accuracy increases. This indicates that the model benefits from the additional feedback of the auxiliary loss during training. 

\setlength{\tabcolsep}{4pt}
\begin{table}[ht]
\begin{center}
\caption{Influence of the loss weights on the overall performance.}
\label{table:ablation_lw}
\begin{tabular}{lC{.8cm}C{.8cm}C{.8cm}C{.8cm}C{.8cm}C{.8cm}C{.8cm}}
\toprule
Loss weight & 0.0 & 0.2 & 0.5 & 0.8 & 1.0 & 2.0 & 3.0 \\
\midrule
Fusion & 94.2 & 92.9 & 93.8 & 94.1 & 93.7 & 92.4 & 93.4 \\
Only Keypoints & 89.2 & 90.8 & 91.2 & 92.1 & 91.0 & 90.1 & 91.7 \\
Only Radar & 84.3 & 81.2 & 83.0 & 84.7 & 82.2 & 81.3 & 82.8 \\

\bottomrule
\end{tabular}
\end{center}
\end{table}
\setlength{\tabcolsep}{1.4pt}

\section{Conclusion}
We present a novel two-stream neural network architecture for the fusion of radar and keypoint data to reliably classify eight different gestures in autonomous driving scenarios. The proposed fusion method improves the classification accuracy over the values obtained with single sensors, while enhancing the recognition robustness in cases of technical sensor failure or adverse environmental conditions. In the model, we first process the data of each modality on its own and then fuse them for the final classification. We propose a stMLP fusion which applies besides the fusion of the features of both modalities also temporal processing. Furthermore, for a better overall performance of our approach, we introduce an auxiliary loss in the training that provides additional feedback to each modality stream. The evaluation of our method on the radar-camera dataset, and we show that even with missing modalities, the model can reach a promising classification performance. In the ablation studies, we demonstrate the influence of the SM-training ratio and the auxiliary loss weight.

\section*{Acknowledgment}
Part of this work was supported by INTUITIVER (7547.223-3/4/), funded by State Ministry of Baden-Württemberg for Sciences, Research and Arts and the State Ministry of Transport Baden-Württemberg.

\clearpage
%
%
\bibliographystyle{splncs04}
\bibliography{references}
\end{document}